\documentclass[sigconf]{acmart}
\AtBeginDocument{%
  \providecommand\BibTeX{{%
    \normalfont B\kern-0.5em{\scshape i\kern-0.25em b}\kern-0.8em\TeX}}}

\usepackage{multirow}
\usepackage{multicol}
\setcopyright{acmcopyright}
\copyrightyear{2023}
\acmYear{2023}
\acmDOI{XXXXXXX.XXXXXXX}

\acmConference[ MMSys]{Make sure to enter the correct
  conference title from your rights confirmation emai}{April 15--18,
  2024}{Bari, Italy}
\acmBooktitle{Bari: ACM Multimedia Systems Conference,
 April 15--18, 2024, Bari, Italy} 
\acmPrice{15.00}
\acmISBN{N/A}

\begin{document}

\title{RecolorCloud: A Point Cloud Tool for Recoloring, Segmentation, and Conversion}

\author{Esteban Segarra Martinez}
\email{esteban.segarra@ucf.edu}
\orcid{1234-5678-9012}
\affiliation{%
  \institution{University of Central Florida}
  \city{Orlando}
  \state{Florida}
  \country{USA}
}

\author{Ryan P. McMahan}
\email{rpm@ucf.edu}
\affiliation{%
  \institution{University of Central Florida}
  % \streetaddress{1 Th{\o}rv{\"a}ld Circle}
  \city{Orlando}
  \state{Florida}
  \country{USA}
  }

\renewcommand{\shortauthors}{Segarra Martinez and McMahan}

\begin{abstract}
Point clouds are a 3D space representation of an environment that was recorded with a high precision laser scanner. These scanners can suffer from environmental interference such as surface shading, texturing, and reflections. Because of this, point clouds may be contaminated with fake or incorrect colors. Current open source or proprietary tools offer limited or no access to correcting these visual errors automatically. 

RecolorCloud is a tool developed to resolve these color conflicts by utilizing automated color recoloring. We offer the ability to deleting or recoloring outlier points automatically with users only needing to specify bounding box regions to effect colors. Results show a vast improvement of the photo-realistic quality of large point clouds. Additionally, users can quickly recolor a point cloud with set semantic segmentation colors. 
\end{abstract}




\maketitle

\section{Introduction}

In recent years, there has been a growing need for high fidelity and photorealisic datasets - also known as laser scans or point cloud data. These datasets are generated using equipment that scan the environment using a laser to determine the distance between the scanner and the environment and combines the positions with a 360 photograph that can provide a color value to the scanned position. 

Point clouds are widely used for high fidelity modeling and rendering, particularly when there is a need to capture a high level of detail from the environment and use that environment for applications such as historical preservation, photorealistic renderings for demonstrations and game environments, and for spatial understanding in robotics. 

More recent applications of point clouds include the Meta Quest series of virtual reality (VR) headsets, which have the capability of scanning the environment and processing the environment in order to determine safe bounds for the user to play in VR and provide game levels if the user wants to play VR inside their point-cloud generated room \cite{Meta23}. Such technologies also allow the possibility of repainting or reskinning the mesh generated from the point cloud and transform the user's point cloud environment into a game-generated environment while preserving the geometry of the user's environment.

With an increasing amount of point cloud datasets there comes a greater need to customize and edit point clouds to fit user's needs. In general, two general categories distinguish an editor, those being if the editor is free and open-source or if the editor is proprietary and close sourced. These categories are further subdivided into editors whose functionality is only to view and have limited editing features and those whose functionality is to be a full suite editor which have features such as registration editing, generating a mesh from the point cloud, and/or can perform some level of automated segmentation. 

In this paper, we present RecolorCloud, a tool that aims to provide a convenient way of recoloring point clouds using semi-automated algorithms, provide a user-controlled method way of segmenting the point cloud, and conversion tool between different point cloud formats. Results from this tool demonstrate that its capabilities to perform recoloring and vastly improve the visual quality of the point cloud. 

This paper will be structured as follows: Section 2 will discuss other point cloud editors and their recoloring capabilities, section 3 will discuss RecolorCloud's features and architecture, section 4 will discuss current and future case studies using the tool, section 5 will discuss the results from the tool and section 6 will conclude the results from RecolorCloud. 

\section{Related Work}

Point cloud editors generally have features that crop, recenter, register, or provide some form of segmenting the point cloud. It is also important to note that not all point cloud editors support recoloring and those that do support recoloring have the feature to create segmentation point clouds, or point clouds that categorize clusters of the point cloud to individual objects. 

\subsection{Editors with Recoloring as Feature}

Because not all recoloring processes are different between editors, we will separate recoloring between \textit{direct recoloring }and \textit{segmentation recoloring}. Direct recoloring refers to the ability of directly recoloring points in order to improve the quality of the point cloud. Segmentation recoloring refers to recoloring the point cloud with high-contrast colors in order to create coarse segmentation categories of objects in the point cloud. 

\begin{table}[]
\centering

\label{table:categories}
\begin{tabular}{|c|c|c|c|}
\hline
\textbf{Editor Name} &
  \textbf{OS?} &
  \textbf{P? } &
  \textbf{Can Recolor?} \\ \hline
   
\begin{tabular}[c]{@{}c@{}}RhinoTerrain \cite{Rhino23}\\  \end{tabular} &
  \begin{tabular}[c]{@{}c@{}}No\\ \end{tabular} &
   Yes &
   Segmentation \\ \hline
   
\begin{tabular}[c]{@{}c@{}}CloudCompare \cite{cloudCompare}\\  \end{tabular} &
  \begin{tabular}[c]{@{}c@{}}Yes\\ \end{tabular} &
   No &
   Direct \\ \hline
   
\begin{tabular}[c]{@{}c@{}}Point Cloud Visualizer \cite{PCVisualizer} \\ \end{tabular} &
  \begin{tabular}[c]{@{}c@{}}No\\ \end{tabular} &
   Yes &
   Direct\\ \hline

\begin{tabular}[c]{@{}c@{}}Vercator Cloud \cite{Vercator23}\\  \end{tabular} &
  \begin{tabular}[c]{@{}c@{}}No\\ \end{tabular} &
   Yes &
   Segmentation \\ \hline
   
\begin{tabular}[c]{@{}c@{}}TCP Point Cloud Editor\cite{TCPPointCloud23}\\  \end{tabular} &
  \begin{tabular}[c]{@{}c@{}}No\\ \end{tabular} &
   Yes &
   Direct \\ \hline

\begin{tabular}[c]{@{}c@{}}Semantic Segmentation Edtr.\cite{hitachi-semantic}\\  \end{tabular} &
  \begin{tabular}[c]{@{}c@{}}Yes\\ \end{tabular} &
   No &
   Segmentation \\ \hline

\begin{tabular}[c]{@{}c@{}}3D BAT\cite{Zimmer_Rangesh_Trivedi_2019}\\  \end{tabular} &
  \begin{tabular}[c]{@{}c@{}}Yes\\ \end{tabular} &
   No &
   No \\ \hline

\begin{tabular}[c]{@{}c@{}}SUSTech Points\cite{SUSTech-POINTS}\\  \end{tabular} &
  \begin{tabular}[c]{@{}c@{}}Yes\\ \end{tabular} &
   No &
   Segmentation \\ \hline

\begin{tabular}[c]{@{}c@{}}Custom Editing Tools \cite{Peteinarelis_2018}\\  \end{tabular} &
  \begin{tabular}[c]{@{}c@{}}No\\ \end{tabular} &
   No &
   Segmentation \\ \hline

\end{tabular}
\caption{Categories of editors and their respective features, Open Source (OS) and Propietary (P)}
\end{table}

As seen in \textbf{Table} \textbf{\ref{table:categories}}, there is an equal amount of tools that support direct and segmentation recoloring. However, a majority of the tools are proprietary and close sourced, with two of the three direct recoloring tools being close sourced. This leaves only CloudCompare\cite{cloudCompare} as the only tool that supports direct recoloring. 

\subsection{Current Open-source Tools}

Not all of the point cloud editors provide large-scale point cloud recoloring. Of the open source tools tested, Point Cloud Visualizer, Semantic Segmentation Editor\cite{hitachi-semantic}, and CloudCompare\cite{cloudCompare} are drastically slow or crashing while performing edits to point clouds that are are larger than 100M points. This limits the capability of editing large point clouds which have a high amount of points for details. 

Open source segmantation tools are common to find, for example there is Point Cloud Labling Tool \cite{behley2019iccv}, which provides a tool for labeling Velodyne data as collected from the KiTTI Dataset. Other tools include Semantic Segmentation Editor\cite{hitachi-semantic}, SUSTech Points \cite{SUSTech-POINTS}, and 3D BAT\cite{Zimmer_Rangesh_Trivedi_2019}. Custom Editing Tools Developed by Alexandros Peteinarelis \cite{Peteinarelis_2018} has support for segmenting point clouds, however, it does not seem to offer a download location for the software discussed in the paper.

Of the discussed editors, Semantic Segmentation Editor\cite{hitachi-semantic} and 3D BAT\cite{Zimmer_Rangesh_Trivedi_2019} provide support for creating bounds based on pre-existing clusters, also known as bounding boxes. Bounds provide coarse selection of points in the point cloud but have the ability of enabling and disabling clusters of the point cloud for consideration. 

\subsection{Impact of Noise on Point Clouds}

There are papers that state the importance of clean or "denoised" point clouds. Noise can impact the subjective quality of the point cloud, as detailed in a study \cite{Zhang_Huang_Zhu_Hwang_2014}  where colored point clouds were subjectively evaluated. Another study \cite{Liu_Su_Duanmu_Liu_Wang_2023} created a dataset of objects with various types of noise introduced by artificial means and corresponding subjective scores by naive subjects. A study \cite{Bolkas_Martinez_2017} observed the impact of color and the effect the shading of glossiness or flatness effects of the estimated distance and color of the objects by two laser scanners. 

In particular, outdoor datasets are prone to have outlier noise produced often by the laser scanner taking geometry from moving objects such as flags or trees and merging incorrect colors from the environment. One paper \cite{Wang_Xu_Liu_Cao_Liu_Yu_Gu_2013} recognized this to be a problem in low quality point clouds and proposed a solution to repair the point cloud geometry introduced from noisy laser scanning. Another paper \cite{Grilli_Menna_Remondino_2017} discussed segmentation and how noise impacted some machine learning algorithms in distinguishing vegetation and building portions. 
%We have to be careful not to go too deep into geometry, we care only about colors. 

Point cloud color noise can also be introduced by interactions such as data file compression. \cite{Freitas_Farias_2021} presents BitDance which is an automated point cloud quality assessor tool which uses factors such as particularly colors and geometry to determine if one point cloud was affected by data compression.

\section{RecolorCloud}

\textit{RecolorCloud} is a recoloring and automated deletion tool that uses bounding boxes for selecting and editing the point cloud. The tool is designed to edit large-scale point clouds in order to edit and delete colors from a point cloud. By using bounds, it is also possible to recolor the point cloud with solid colors or split the point cloud into individualized point clouds. As an additional feature, RecolorCloud provides convenient file conversion between different popular point cloud formats. 

\textit{RecolorCloud} provides the following features:
\begin{itemize}
    \item Large point cloud support (>100 million points)
    \item Recoloring and deleting points based on coloring criteria including segmentation
    \item File conversion between popular file formats
    \item Fragmentation of point cloud into smaller ones based on bounds
    \item Open source and free of charge
\end{itemize}

The tool was created to be as easy to use for a researcher and novice user in point cloud editing. The computational cost of running this tool is to have enough RAM to load the point cloud in consideration. This project was completed and tested using a computer with an Intel 10850k, 128GB of DDR4 RAM, and a Nvidia GTX 2080. 

\subsection{Implementation and Architecture}

\begin{figure}[!h]
    \centering
    \includegraphics[width=.35\textwidth]{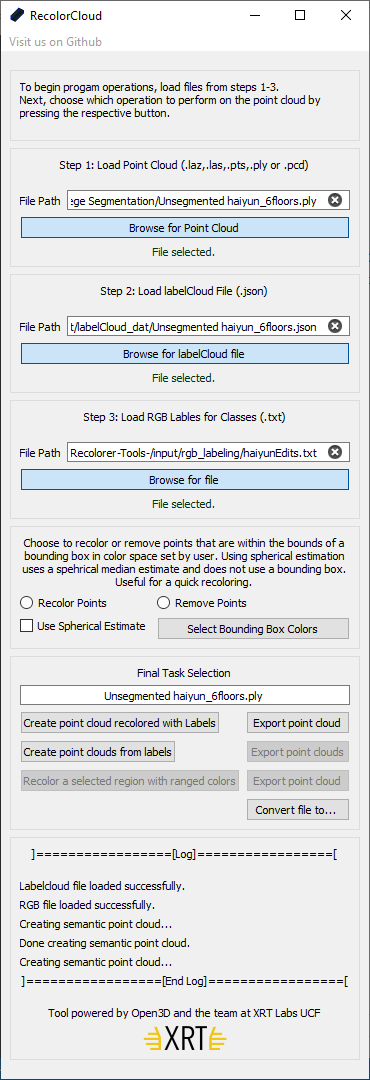}
    \caption{User Interface of the RecolorCloud Software.}
    \label{fig:softwareUI}
\end{figure}

This project was implemented using Python 3.6.3 with a Anaconda environment for dependency control. \textbf{Figure \ref{fig:softwareUI}} shows the user interface that displays RecolorCloud's functions. The application runs standalone, however it requires input data representing the bounding boxes that are meant to be edited. Currently we provide those bouding boxes using another Python tool called LabelCloud \cite{Sager_Zschech_Kuhl_2021}. LabelCloud opens the point cloud and provides a convenient interface for creating and placing bounding boxes. The python packages used include Open3D\cite{Zhou2018}, SciPy \cite{2020SciPy-NMeth}, Numpy\cite{harris2020array}, and PyQt5. The dependency list was explicitly kept short so as to have the least amount of extra pacakges installed.  

\subsection{Usage}

For all cases, there has to be a point cloud loaded to edit the point cloud. These are for cases when the point cloud needs to be converted between file types. If direct or semantically recoloring the point cloud, a LabelCloud bounding file and a custom text file are used to control the semantic colors and enable the different labeled bounds. 

After loading the three files, the user has the option of semantically recoloring the bounded regions, deleting regions of the point cloud, recoloring them using criteria, or splitting the point cloud into smaller ones. 

% - Python plus Open3D etc.
% - Including an architecture diagram

\subsection{Feature 1: Recoloring points}
\label{sec:feature1}
Recoloring points is done by the user selecting the option in the UI to recolor the points in the bounding boxes. Recoloring done by \textit{RecolorCloud} primarily works in the RGB 3D space. The user then can select between two different recoloring algorithms as described in the following subsections:

\subsubsection{Spherical Recoloring}
\label{sec:sphericalRecoloring}
This technique utilizes the color space from a bounding box, finds the mean color point from the color space, and generates a sphere. All points inside the sphere are preserved while points outside the sphere are recolored with points inside the sphere. This technique is optimal for finding the average color from a region, such as a tree, pole, or bushes. 

\subsubsection{RGB Bounding Box}
\label{sec:RGBBBox}
This technique takes the color space from all unique colors inside a bounding box and generates a bounding box in the RGB space and corresponding centroid. The user defines a new bounding box region which is used to move the centroid of the first bounding box and scale it down to the size of the user-defined bounding box. This technique is optimal for changing the color of a selected region while retaining the brightness and shading of points inside the selected bounding box. 

\subsubsection{RGB Color Substitution}

This is a direct color replacement of colors based on custom file that define colors for each bounding box. This is useful for recoloring entire portions of a point cloud into a semantically segmented point cloud. This technique only focuses on substituting the colors inside the bounding boxes and recolors them with user-defined colors. Points not inside the bounding boxes are not recolored and removed from the final point cloud. 

\subsection{Feature 2: Deleting outlier colors}
\label{sec:feature2}
This feature utilizes the same algorithms as outlined in subsections \ref{sec:sphericalRecoloring} and \ref{sec:RGBBBox}.Functionally however, the way outlier points are handled is different from the recoloring algorithms. The changed behavior of the algorithms are changed as follow: 

\begin{itemize}
    \item Spherical Recoloring: Points that lie outside the sphere from  are deleted rather than recolored. 
    \item RGB Bounding Box: Similarly, points that lie outside the user-defined domain from the technique are deleted.
\end{itemize}

\subsection{Feature 3: Splitting the point cloud}
\label{sec:feature3}
This technique splits up the point cloud into individual components as defined by bounding boxes. This feature is useful for cases such as a large-scale point and there's a need to break down the point cloud into individually labeled point cloud objects. 

\subsection{Feature 4: Point cloud conversion}
\label{sec:feature4}
An additional feature provided by RecolorCloud is the ability to convert point clouds between different formats. The application can load and save files in the following formats: .las, .laz, .xyz, .xyzn, .xyzrgb, .pts, .ply, and .pcd. This allows users to efficiently convert large point clouds and export them quickly without crashing the application due to visualization or unsupported file types. 

\section{Case Studies}

RecolorCloud has been applied successfully applied on different popular datasets and case studies. Below are the results obtained from different case applications. For the purposes of consistency in previewing the results, the before and after images of the point clouds are shown using Autodesk Recap \cite{Recap23} with a solid black background. 

\subsection{Greek Park Dataset Case Study}

This is a dataset that was collected with the help of the University of Central Florida's (UCF) ChronoPoints laser scanning project\cite{Chrono23}. ChronoPoints provided a dataset collected in-situ of UCF's Greek Park fraternity road and buildings. The dataset is approximately 192 meters long by 172 meters wide, providing access to environmental elements such as trees and buildings. This dataset contained multiple noisy outliers within its trees. Additionally, a semantically segmented version of this dataset was generated.

\begin{figure}[!h]
    \centering
    \includegraphics[width=.5\textwidth]{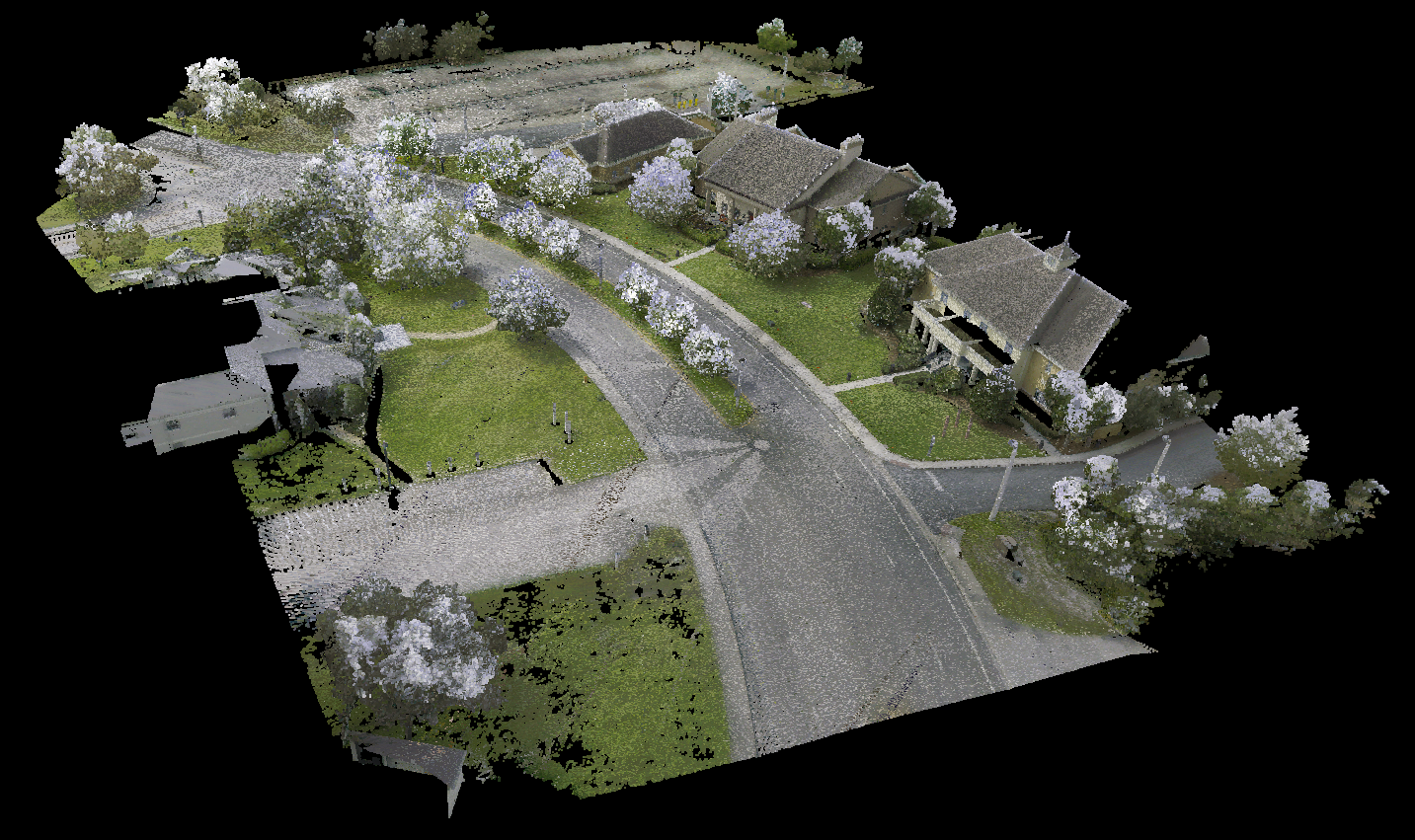}
    % \caption{Uncorrected Greek Park Point Cloud.}
%     \label{fig:unfixedGP}
% \end{figure}
% \begin{figure}[!h]
%     \centering
    \includegraphics[width=.5\textwidth]{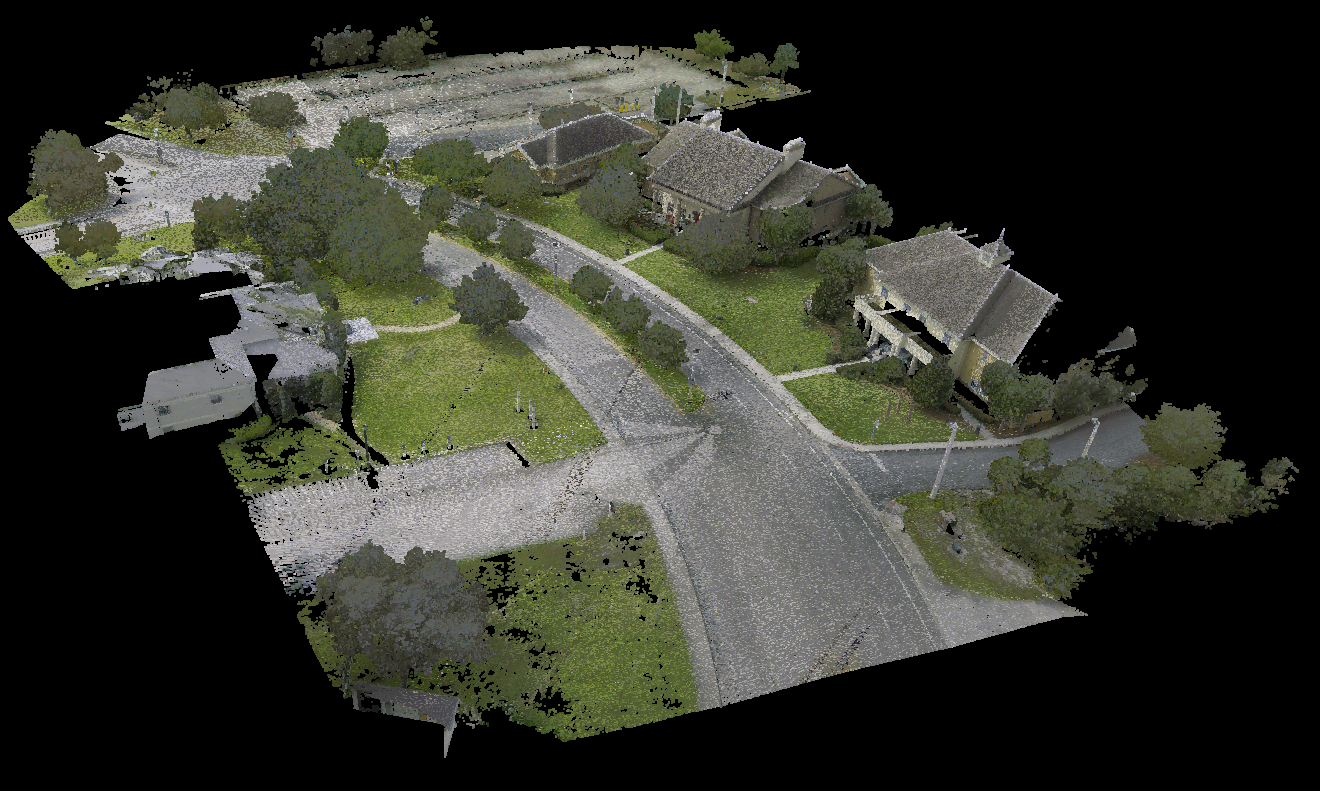}
    \caption{Greek Park Point Cloud Uncorrected (Top Panel) and Corrected (Bottom Panel).}
    \label{fig:fixedGP}
\end{figure}

\begin{figure}[!h]
    \centering
    \includegraphics[width=.5\textwidth]{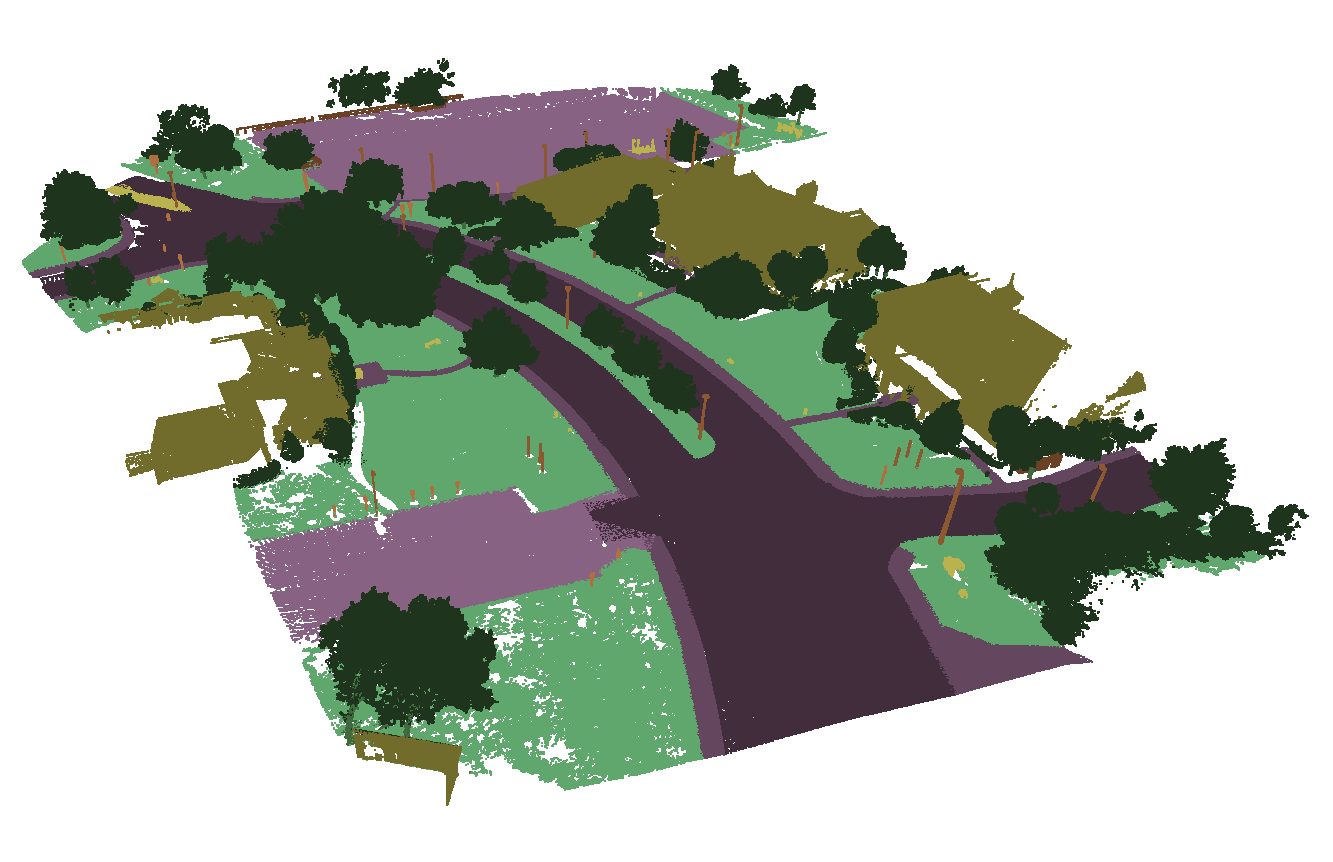}
    \caption{Semantic Coloring of Greek Park Point Cloud.}
    \label{fig:semanticGP}
\end{figure}
% - Images of improvements

% - Semantic representations

\subsubsection{Recoloring Outlier Points}
This dataset had issues with almost all the trees as seen in \textbf{figure \ref{fig:fixedGP}}. As this was an outdoor dataset, the trees got contaminated because the wind moved the leaves which were registered with laser scanner and then got merged with the photographs captured by the scanner. The errors manifested as white points with a combination of blue-green points from the color of the sky. 

These errors were corrected by a combination of removing excess white outlier points and recoloring the remaining outlier blue-green points to be closer to the color of the tree. The results from this process can be seen in \textbf{figure \ref{fig:fixedGP}}. 

\subsubsection{Semantic Segmentation of Greek Park}
Using the same bounding boxes from the recoloring task, we recolored the point cloud using the KiTTI dataset's colors. The advantage of doing this process is that the point cloud can then be rendered using a synthetic camera. The results from the segmentation process in \textbf{figure \ref{fig:semanticGP}}.

%need to cite kitti here. 
\subsection{De-noising and Recoloring Case Study}

A second case that RecolorCloud has been tested on has been the use of the dataset Tanks and Temples\cite{Knapitsch_Park_Zhou_Koltun_2017}, which features a selection of indoor and outdoor datasets. In particular, there was an outdoor dataset called Barn which featured noisy outlier errors commonly associated with outdoor datasets. This was done to demonstrate tool's capability of repairing the noisy colors from the dataset. Below are the results from running RecolorCloud on this case study.

% - Images of improvements

\begin{figure}[!h]
    \centering
    \includegraphics[width=.5\textwidth]{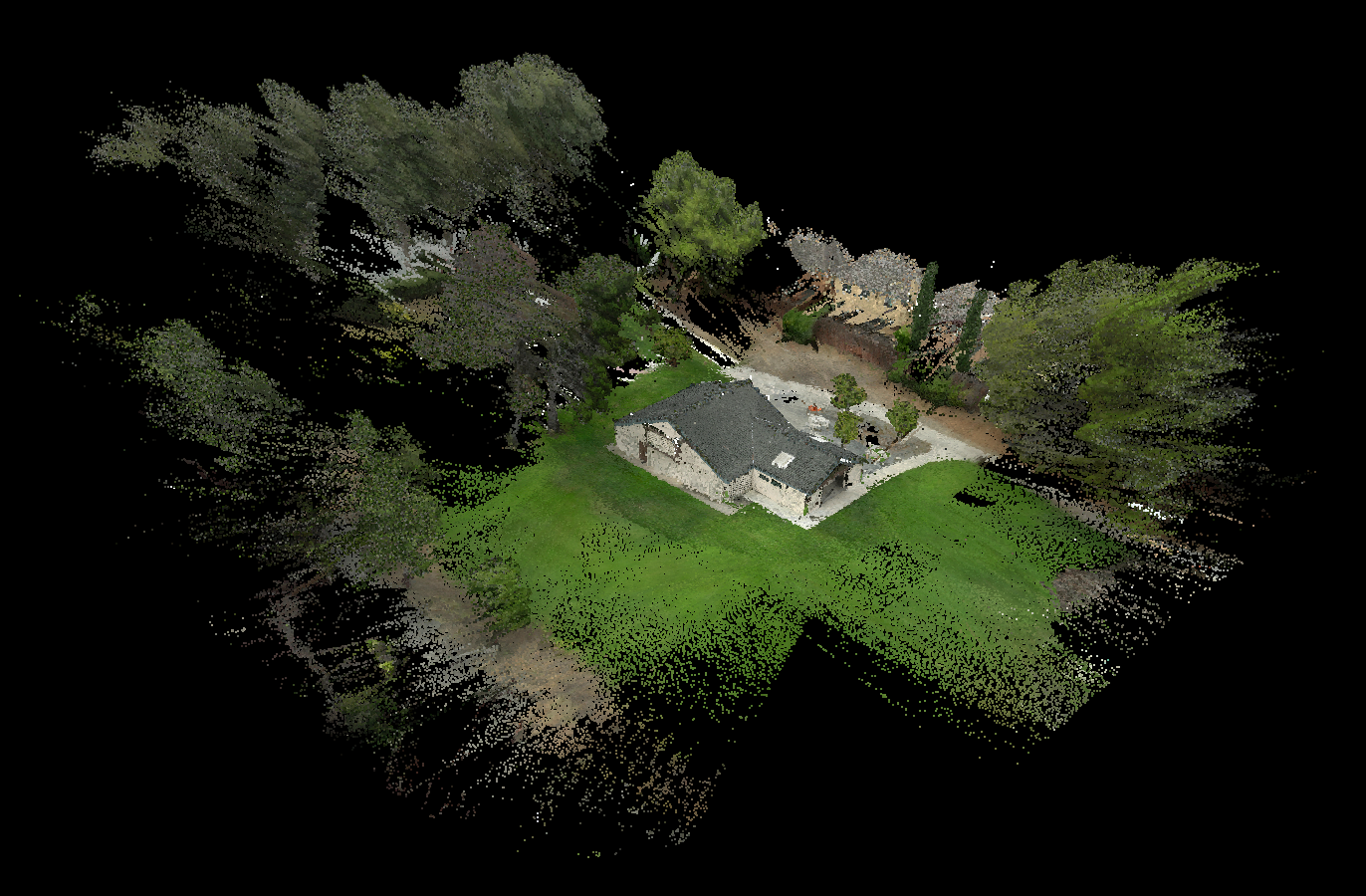}
    % \caption{Uncorrected Barn Point Cloud.}
    % \label{fig:unfixedBarn}
% \end{figure}
% \begin{figure}[!h]
%     \centering
    \includegraphics[width=.5\textwidth]{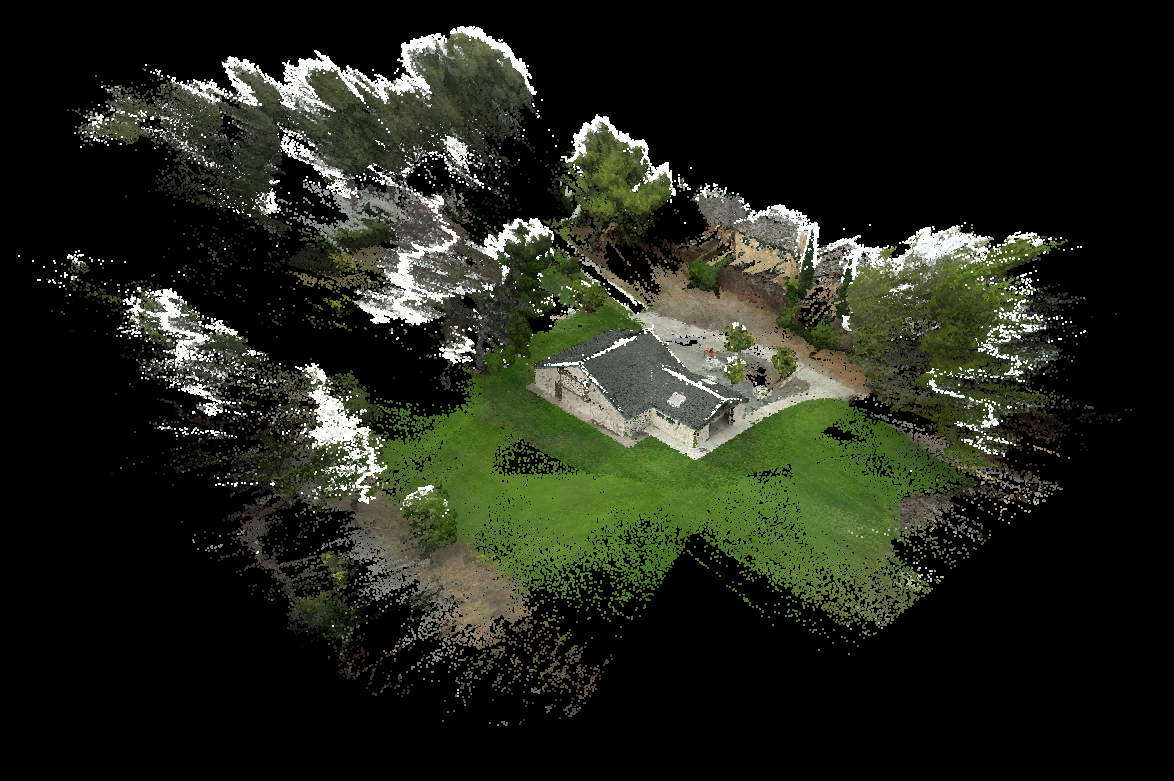}
    \caption{Barn Point Cloud Uncorrected (Top Panel) and Corrected (Bottom Panel).}
    \label{fig:fixedBarn}
\end{figure}

\subsubsection{Color Outlier Correction} 
Similar to the Greek Park dataset, the trees and the buildings had outlier errors in the form of white points as seen on the top panel of \textbf{figure \ref{fig:fixedBarn}}. This dataset was corrected by placing bounding boxes on the trees and the buildings and applying the spherical recoloring technique as seen on the bottom panel of \textbf{figure \ref{fig:fixedBarn}}. 
% The end-results are displayed in \textbf{figure \ref{fig:fixedBarn}}. 

\begin{figure}[!h]
    \centering
    \includegraphics[width=.5\textwidth]{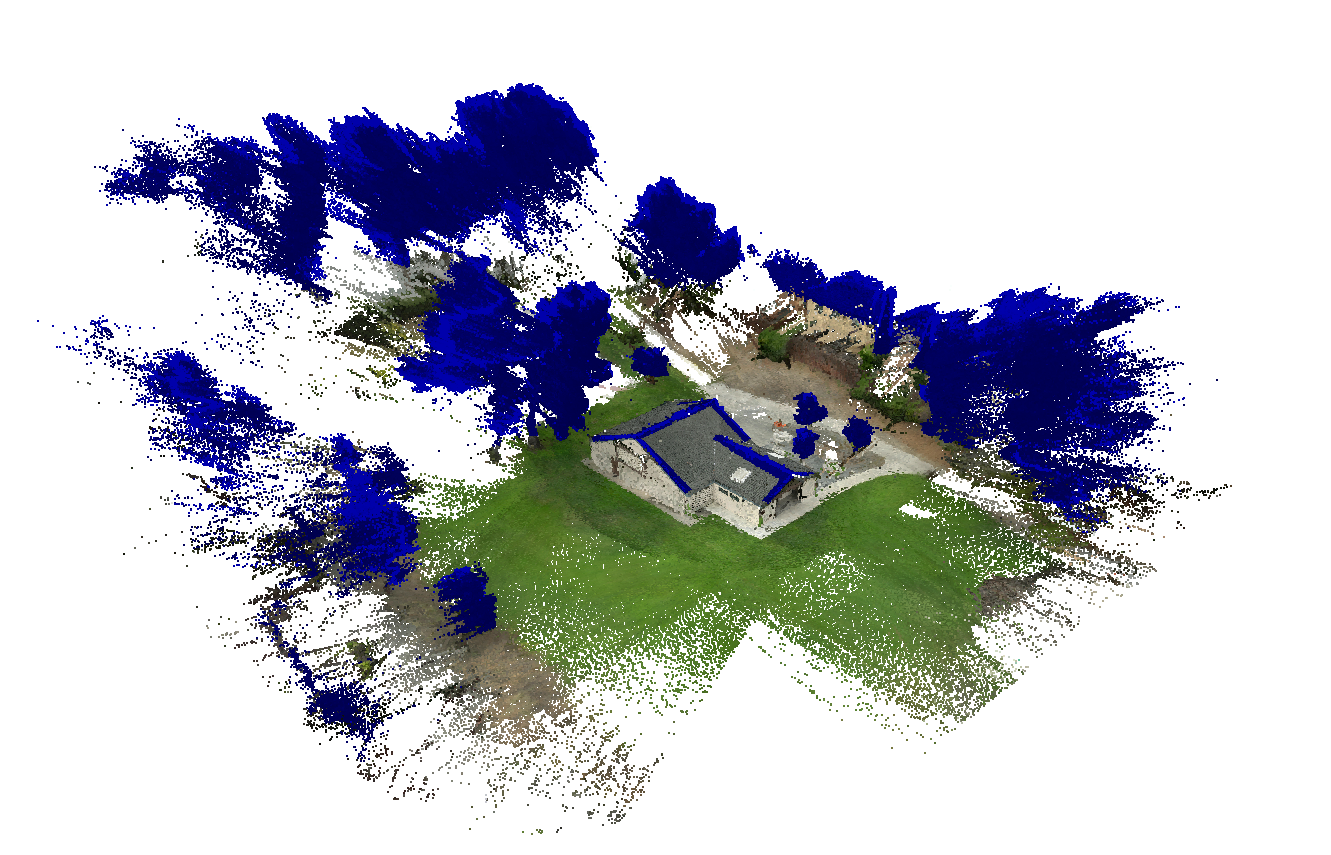}
    \caption{Color Shifted Barn Point Cloud.}
    \label{fig:hueshiftBarn}
\end{figure}

\subsubsection{Color Shift Application}

As part of the tool's ability to recolor, the color of items in bounding boxes can be color-shifted to other hues as stated in section \ref{sec:RGBBBox}. In the example of the barn dataset, the  color of the trees and the corrections on the building were color shifted to deep blue to indicate the potential of color shifting entire sections of the point cloud. This is different from semantic recoloring as colors are not of one unique color and are rather shifted colors from the original set. 

\subsection{Semantic and Hue Recoloring Case Study}

RecolorCloud was applied to an indoor dataset, the Multisensor Indoor Mapping and Positioning Dataset \cite{Wang_Hou_Wen_Gong_Li_Sun_Li_2018}. This is to demonstrate the capabilities of the tool to semantically recolor the dataset. The point cloud chosen from this collection was the \textit{Colored Indoor Laser Scanning Dataset}, named \textit{20180526haiyun6floors.ply}.  This dataset was separately labeled with bounding boxes by LabelCloud and colored with custom segmentation colors. 

% - Images of improvements

\begin{figure}[!h]
    \centering
    \includegraphics[width=.5\textwidth]{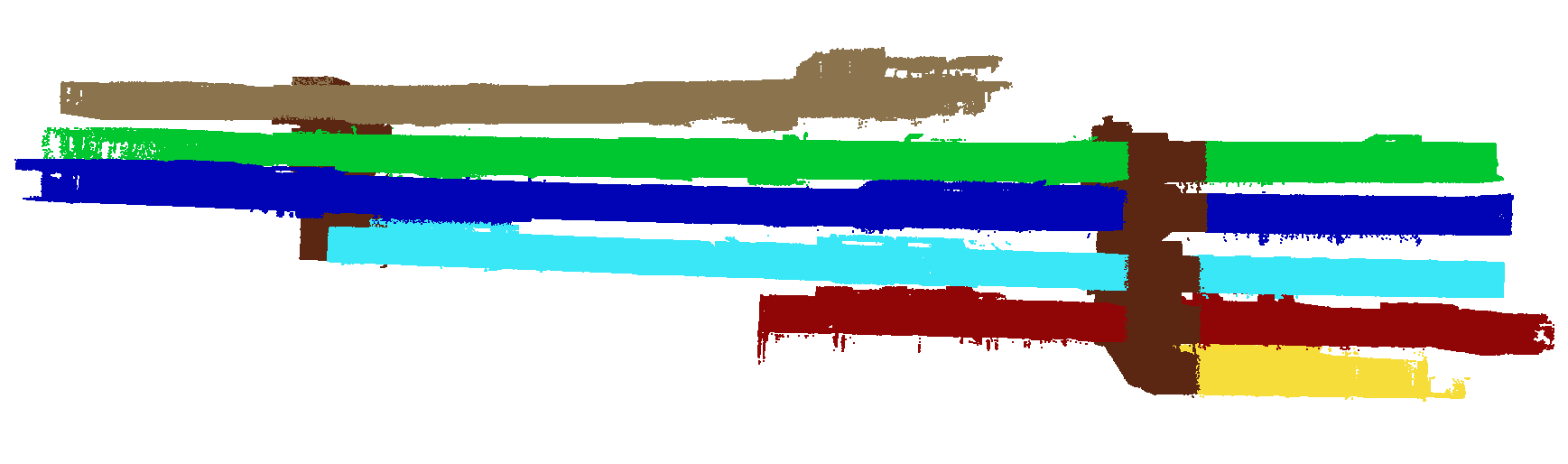}
    % \caption{Unsegmented Point Cloud.}
%     \label{fig:unsegmentedHaiyun}
% \end{figure}
% \begin{figure}[!h]
%     \centering
    \includegraphics[width=.5\textwidth]{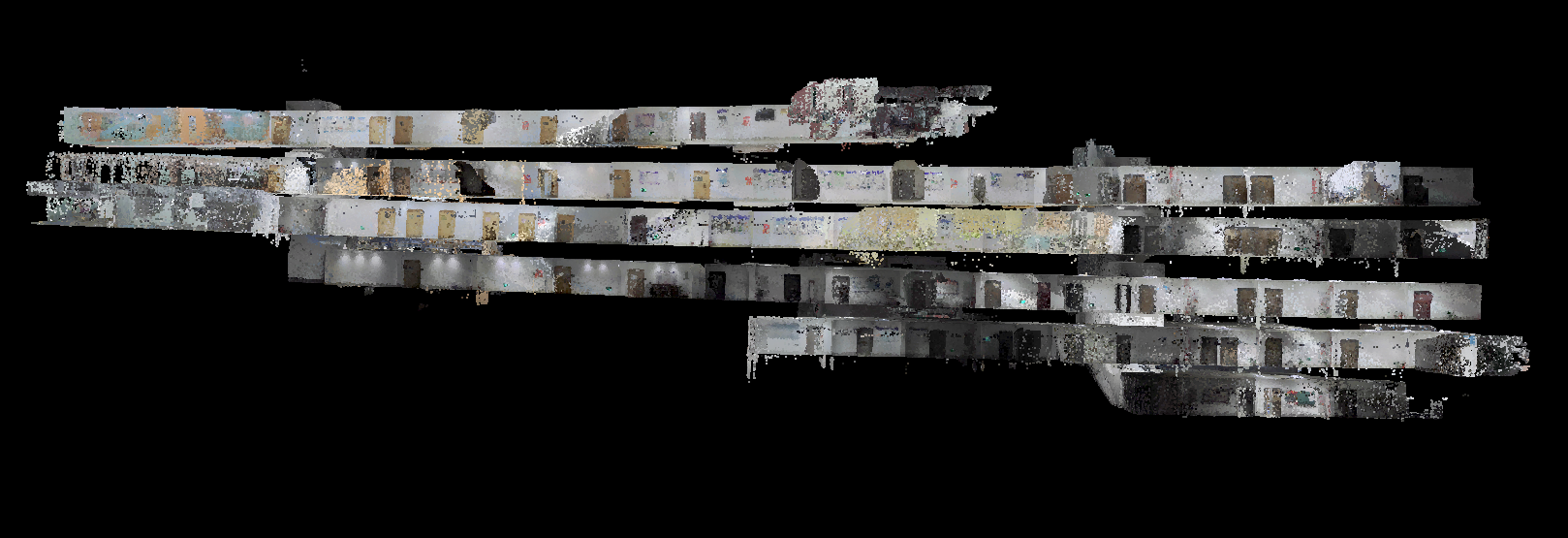}
    \caption{Segmented Point Cloud.}
    \label{fig:segmentedHaiyun}
\end{figure}

This dataset had multiple outlier points from rooms that were partially scanned by the mobile laser scanner. These points were removed before proceeding with the recoloring with a preview seen in the top panel of \textbf{figure \ref{fig:segmentedHaiyun}}. After applying the bounding boxes on the point cloud the final result of semantically segmenting the point cloud can be seen on the bottom panel of\textbf{figure \ref{fig:segmentedHaiyun}}.

\section{Discussion}

The tool has proven its capabilities in being able to perform direct and semantic recoloring. The following sections describes the tool's capabilities, current limitations, and future planned features. 

\subsection{New Capabilities} 

The greatest highlight in RecolorCloud's capability is its ability to handle large scale point clouds and efficiently recolor them or delete points based on desired criteria. In addition, RecolorCloud is designed such that additional features can be easily coded into the application. This way, if someone wanted a lightweight semantic or direct recoloring tool, they will be able to download the tool and apply the changes easily. 

\subsection{Limitations}

First, the application depends on LabelCloud for generating bounding boxes. This limitation means that RecolorCloud can only perform coarse selection and editing of the point cloud. Granular selection will remain a future work element that will be corrected with improvements such as lasso selection.

Second, the user interface does not directly display the changes that will be applied to the point cloud before editing. This is because to preview the edits done on a large-scale point cloud will require performing the edits, which is a slow process. 

Third, this application depends on having a python back-end for running. This creates a limitation for novice users to use the software as they would have to install Conda, install related dependencies, and run RecolorCloud on a console window. Additionally, users have to install and run LabelCloud to generate the bounding boxes to select points on the point cloud - which additionally requires a down-sampled point cloud to simply view the point cloud in LabelCloud without performance issues. 

\subsection{Future Work}

Future work will make RecolorCloud a PyPi package for ease of installation and distribution. Additionally, on the Github distribution of the source code we plan to discover and implement recommendations by the research community. A future version of this tool is currently under development which will remove the dependency of using LabelCloud for generating bounding boxes, incorporate point cloud selection, and a improved user interface.  

\section{Conclusion}

Point clouds will remain an important data format in the future and will require tools to be able to manipulate and correct them. There are a few tools that are capable of editing the colors of point clouds directly, however not all are free or open source. Current open source solutions provide limited capabilities of editing colors with CloudCompare being an exception but limited to medium sized point clouds. Tools that provide a dedicated segmentation mechanic, such as 3D BAT, Semantic Segmentation Editor, or SUSTech Points, can create segmented boundaries but are focused on labeling points rather than recoloring the point cloud and crash when opening large point clouds. 

\textbf{RecolorCloud}, aims to solve the issue of a lack of a tool that provides users the ability to recolor point clouds directly and correct point clouds. By providing researchers with an open-source python application, users have an accessible option to perform edits and corrections to their point clouds. From the results, RecolorCloud is capable of recoloring by segmentation, correcting outlier colors errors, segmenting per bounding box, and converting between multiple popular point cloud formats. 

\bibliographystyle{ACM-Reference-Format}
\bibliography{references}


\begin{thebibliography}{25}

%%% the Web address.

\ifx \showCODEN    \undefined \def \showCODEN     #1{\unskip}     \fi
\ifx \showDOI      \undefined \def \showDOI       #1{#1}\fi
\ifx \showISBNx    \undefined \def \showISBNx     #1{\unskip}     \fi
\ifx \showISBNxiii \undefined \def \showISBNxiii  #1{\unskip}     \fi
\ifx \showISSN     \undefined \def \showISSN      #1{\unskip}     \fi
\ifx \showLCCN     \undefined \def \showLCCN      #1{\unskip}     \fi
\ifx \shownote     \undefined \def \shownote      #1{#1}          \fi
\ifx \showarticletitle \undefined \def \showarticletitle #1{#1}   \fi
\ifx \showURL      \undefined \def \showURL       {\relax}        \fi
% invisible to TeX
\providecommand\bibfield[2]{#2}
\providecommand\bibinfo[2]{#2}
\providecommand\natexlab[1]{#1}
\providecommand\showeprint[2][]{arXiv:#2}

\bibitem[Andrija(2023)]%
        {TCPPointCloud23}
\bibfield{author}{\bibinfo{person}{Andrija}.} \bibinfo{year}{2023}\natexlab{}.
\newblock \bibinfo{booktitle}{\emph{TCP Point Cloud Editor}}.
\newblock
\urldef\tempurl%
\url{https://www.adriabim.com/tcp-pointcloud-editor-edit-view-and-manage-point-clouds/}
\showURL{%
Retrieved October 18, 2023 from \tempurl}


\bibitem[Behley et~al\mbox{.}(2019)]%
        {behley2019iccv}
\bibfield{author}{\bibinfo{person}{J. Behley}, \bibinfo{person}{M. Garbade}, \bibinfo{person}{A. Milioto}, \bibinfo{person}{J. Quenzel}, \bibinfo{person}{S. Behnke}, \bibinfo{person}{C. Stachniss}, {and} \bibinfo{person}{J. Gall}.} \bibinfo{year}{2019}\natexlab{}.
\newblock \showarticletitle{{SemanticKITTI: A Dataset for Semantic Scene Understanding of LiDAR Sequences}}. In \bibinfo{booktitle}{\emph{Proc. of the IEEE/CVF International Conf.~on Computer Vision (ICCV)}}.
\newblock


\bibitem[Bolkas and Martinez(2017)]%
        {Bolkas_Martinez_2017}
\bibfield{author}{\bibinfo{person}{Dimitrios Bolkas} {and} \bibinfo{person}{Aaron Martinez}.} \bibinfo{year}{2017}\natexlab{}.
\newblock \showarticletitle{Effect of target color and scanning geometry on terrestrial LiDAR point-cloud noise and plane fitting}.
\newblock \bibinfo{journal}{\emph{Journal of Applied Geodesy}}  \bibinfo{volume}{12} (\bibinfo{date}{Dec.} \bibinfo{year}{2017}).
\newblock
\urldef\tempurl%
\url{https://doi.org/10.1515/jag-2017-0034}
\showDOI{\tempurl}


\bibitem[Diniz et~al\mbox{.}(2021)]%
        {Freitas_Farias_2021}
\bibfield{author}{\bibinfo{person}{Rafael Diniz}, \bibinfo{person}{Pedro Garcia~Freitas}, {and} \bibinfo{person}{Mylene Farias}.} \bibinfo{year}{2021}\natexlab{}.
\newblock \showarticletitle{Color and Geometry Texture Descriptors for Point-Cloud Quality Assessment}.
\newblock \bibinfo{journal}{\emph{IEEE Signal Processing Letters}}  \bibinfo{volume}{28} (\bibinfo{year}{2021}), \bibinfo{pages}{1150–1154}.
\newblock
\showISSN{1070-9908, 1558-2361}
\urldef\tempurl%
\url{https://doi.org/10.1109/LSP.2021.3088059}
\showDOI{\tempurl}


\bibitem[Grilli et~al\mbox{.}(2017)]%
        {Grilli_Menna_Remondino_2017}
\bibfield{author}{\bibinfo{person}{E. Grilli}, \bibinfo{person}{F. Menna}, {and} \bibinfo{person}{F. Remondino}.} \bibinfo{year}{2017}\natexlab{}.
\newblock \showarticletitle{A Review of Point Clouds Segmentation and Classification Algorithms}.
\newblock \bibinfo{journal}{\emph{The International Archives of the Photogrammetry, Remote Sensing and Spatial Information Sciences}}  \bibinfo{volume}{XLII-2/W3} (\bibinfo{date}{Feb.} \bibinfo{year}{2017}), \bibinfo{pages}{339–344}.
\newblock
\showISSN{2194-9034}
\urldef\tempurl%
\url{https://doi.org/10.5194/isprs-archives-XLII-2-W3-339-2017}
\showDOI{\tempurl}


\bibitem[Harris et~al\mbox{.}(2020)]%
        {harris2020array}
\bibfield{author}{\bibinfo{person}{Charles~R. Harris}, \bibinfo{person}{K.~Jarrod Millman}, \bibinfo{person}{St{\'{e}}fan~J. van~der Walt}, \bibinfo{person}{Ralf Gommers}, \bibinfo{person}{Pauli Virtanen}, \bibinfo{person}{David Cournapeau}, \bibinfo{person}{Eric Wieser}, \bibinfo{person}{Julian Taylor}, \bibinfo{person}{Sebastian Berg}, \bibinfo{person}{Nathaniel~J. Smith}, \bibinfo{person}{Robert Kern}, \bibinfo{person}{Matti Picus}, \bibinfo{person}{Stephan Hoyer}, \bibinfo{person}{Marten~H. van Kerkwijk}, \bibinfo{person}{Matthew Brett}, \bibinfo{person}{Allan Haldane}, \bibinfo{person}{Jaime~Fern{\'{a}}ndez del R{\'{i}}o}, \bibinfo{person}{Mark Wiebe}, \bibinfo{person}{Pearu Peterson}, \bibinfo{person}{Pierre G{\'{e}}rard-Marchant}, \bibinfo{person}{Kevin Sheppard}, \bibinfo{person}{Tyler Reddy}, \bibinfo{person}{Warren Weckesser}, \bibinfo{person}{Hameer Abbasi}, \bibinfo{person}{Christoph Gohlke}, {and} \bibinfo{person}{Travis~E. Oliphant}.} \bibinfo{year}{2020}\natexlab{}.
\newblock \showarticletitle{Array programming with {NumPy}}.
\newblock \bibinfo{journal}{\emph{Nature}} \bibinfo{volume}{585}, \bibinfo{number}{7825} (\bibinfo{date}{Sept.} \bibinfo{year}{2020}), \bibinfo{pages}{357--362}.
\newblock
\urldef\tempurl%
\url{https://doi.org/10.1038/s41586-020-2649-2}
\showDOI{\tempurl}


\bibitem[Knapitsch et~al\mbox{.}(2017)]%
        {Knapitsch_Park_Zhou_Koltun_2017}
\bibfield{author}{\bibinfo{person}{Arno Knapitsch}, \bibinfo{person}{Jaesik Park}, \bibinfo{person}{Qian-Yi Zhou}, {and} \bibinfo{person}{Vladlen Koltun}.} \bibinfo{year}{2017}\natexlab{}.
\newblock \showarticletitle{Tanks and temples: benchmarking large-scale scene reconstruction}.
\newblock \bibinfo{journal}{\emph{ACM Transactions on Graphics}} \bibinfo{volume}{36}, \bibinfo{number}{4} (\bibinfo{date}{Aug.} \bibinfo{year}{2017}), \bibinfo{pages}{1–13}.
\newblock
\showISSN{0730-0301, 1557-7368}
\urldef\tempurl%
\url{https://doi.org/10.1145/3072959.3073599}
\showDOI{\tempurl}


\bibitem[Li et~al\mbox{.}(2020)]%
        {SUSTech-POINTS}
\bibfield{author}{\bibinfo{person}{E Li}, \bibinfo{person}{Shuaijun Wang}, \bibinfo{person}{Chengyang Li}, \bibinfo{person}{Dachuan Li}, \bibinfo{person}{Xiangbin Wu}, {and} \bibinfo{person}{Qi Hao}.} \bibinfo{year}{2020}\natexlab{}.
\newblock \showarticletitle{SUSTech POINTS: A Portable 3D Point Cloud Interactive Annotation Platform System}. In \bibinfo{booktitle}{\emph{2020 IEEE Intelligent Vehicles Symposium (IV)}}. \bibinfo{pages}{1108--1115}.
\newblock
\urldef\tempurl%
\url{https://doi.org/10.1109/IV47402.2020.9304562}
\showDOI{\tempurl}


\bibitem[Liu et~al\mbox{.}(2023)]%
        {Liu_Su_Duanmu_Liu_Wang_2023}
\bibfield{author}{\bibinfo{person}{Qi Liu}, \bibinfo{person}{Honglei Su}, \bibinfo{person}{Zhengfang Duanmu}, \bibinfo{person}{Wentao Liu}, {and} \bibinfo{person}{Zhou Wang}.} \bibinfo{year}{2023}\natexlab{}.
\newblock \showarticletitle{Perceptual Quality Assessment of Colored 3D Point Clouds}.
\newblock \bibinfo{journal}{\emph{IEEE Transactions on Visualization and Computer Graphics}} \bibinfo{volume}{29}, \bibinfo{number}{8} (\bibinfo{date}{Aug.} \bibinfo{year}{2023}), \bibinfo{pages}{3642–3655}.
\newblock
\showISSN{1077-2626, 1941-0506, 2160-9306}
\urldef\tempurl%
\url{https://doi.org/10.1109/TVCG.2022.3167151}
\showDOI{\tempurl}


\bibitem[Mandrioli et~al\mbox{.}({[n.\,d.]})]%
        {hitachi-semantic}
\bibfield{author}{\bibinfo{person}{Damien Mandrioli}, \bibinfo{person}{Wei Mingzhi}, \bibinfo{person}{Dario Široki}, \bibinfo{person}{Daniele Morabito}, {and} \bibinfo{person}{Bruce Chou}.} \bibinfo{year}{[n.\,d.]}\natexlab{}.
\newblock \bibinfo{booktitle}{\emph{Semantic Segmentation Editor}}.
\newblock
\urldef\tempurl%
\url{https://www.scilab.org/}
\showURL{%
\tempurl}


\bibitem[N/A(2023a)]%
        {Recap23}
\bibfield{author}{\bibinfo{person}{N/A}.} \bibinfo{year}{2023}\natexlab{a}.
\newblock \bibinfo{booktitle}{\emph{Autodesk Recap Features}}.
\newblock
\urldef\tempurl%
\url{https://www.autodesk.com/products/recap/features}
\showURL{%
Retrieved October 18, 2023 from \tempurl}


\bibitem[N/A(2023b)]%
        {Chrono23}
\bibfield{author}{\bibinfo{person}{N/A}.} \bibinfo{year}{2023}\natexlab{b}.
\newblock \bibinfo{booktitle}{\emph{ChronoPoints: Documenting the Life and Culture of Communities Through Structures, Stories, and Artifacts}}.
\newblock
\urldef\tempurl%
\url{https://chronopoints.eecs.ucf.edu/about/}
\showURL{%
Retrieved October 18, 2023 from \tempurl}


\bibitem[N/A(2023c)]%
        {cloudCompare}
\bibfield{author}{\bibinfo{person}{N/A}.} \bibinfo{year}{2023}\natexlab{c}.
\newblock \bibinfo{booktitle}{\emph{CloudCompare: 3D point cloud and mesh processing software Open Source Project}}.
\newblock
\urldef\tempurl%
\url{https://www.danielgm.net/cc/}
\showURL{%
Retrieved October 18, 2023 from \tempurl}


\bibitem[N/A(2023d)]%
        {Meta23}
\bibfield{author}{\bibinfo{person}{N/A}.} \bibinfo{year}{2023}\natexlab{d}.
\newblock \bibinfo{booktitle}{\emph{Learn about point clouds with Meta Quest}}.
\newblock
\urldef\tempurl%
\url{https://www.meta.com/help/quest/articles/in-vr-experiences/oculus-features/point-cloud/}
\showURL{%
Retrieved October 16, 2023 from \tempurl}


\bibitem[N/A(2023e)]%
        {Vercator23}
\bibfield{author}{\bibinfo{person}{N/A}.} \bibinfo{year}{2023}\natexlab{e}.
\newblock \bibinfo{booktitle}{\emph{Registration Service}}.
\newblock
\urldef\tempurl%
\url{https://vercator.com/register-laser-scans-in-the-cloud/}
\showURL{%
Retrieved October 18, 2023 from \tempurl}


\bibitem[N/A(2023f)]%
        {Rhino23}
\bibfield{author}{\bibinfo{person}{N/A}.} \bibinfo{year}{2023}\natexlab{f}.
\newblock \bibinfo{booktitle}{\emph{RhinoTerrain}}.
\newblock
\urldef\tempurl%
\url{https://www.rhinoterrain.com/en/rhinoterrain.html}
\showURL{%
Retrieved October 18, 2023 from \tempurl}


\bibitem[Peteinarelis(2018)]%
        {Peteinarelis_2018}
\bibfield{author}{\bibinfo{person}{Alexandros Peteinarelis}.} \bibinfo{year}{2018}\natexlab{}.
\newblock \showarticletitle{Custom point cloud edit and analysis tools in visual programming: Evaluation of heritage facades}. In \bibinfo{booktitle}{\emph{2018 3rd Digital Heritage International Congress (DigitalHERITAGE) held jointly with 2018 24th International Conference on Virtual Systems I\& Multimedia (VSMM 2018)}}. \bibinfo{publisher}{IEEE}, \bibinfo{address}{San Francisco, CA, USA}, \bibinfo{pages}{1–7}.
\newblock
\showISBNx{978-1-72810-292-4}
\urldef\tempurl%
\url{https://doi.org/10.1109/DigitalHeritage.2018.8810052}
\showDOI{\tempurl}


\bibitem[Sager et~al\mbox{.}(2021)]%
        {Sager_Zschech_Kuhl_2021}
\bibfield{author}{\bibinfo{person}{Christoph Sager}, \bibinfo{person}{Patrick Zschech}, {and} \bibinfo{person}{Niklas Kuhl}.} \bibinfo{year}{2021}\natexlab{}.
\newblock \showarticletitle{labelCloud: A Lightweight Domain-Independent Labeling Tool for 3D Object Detection in Point Clouds}. In \bibinfo{booktitle}{\emph{CAD’21 Proceedings}}. \bibinfo{publisher}{CAD Solutions LLC}, \bibinfo{pages}{319–323}.
\newblock
\urldef\tempurl%
\url{https://doi.org/10.14733/cadconfP.2021.319-323}
\showDOI{\tempurl}


\bibitem[Uhlík({[n.\,d.]})]%
        {PCVisualizer}
\bibfield{author}{\bibinfo{person}{Jakub Uhlík}.} \bibinfo{year}{[n.\,d.]}\natexlab{}.
\newblock \bibinfo{booktitle}{\emph{Point Cloud Visualizer}}.
\newblock
\urldef\tempurl%
\url{https://blendermarket.com/products/pcv}
\showURL{%
\tempurl}


\bibitem[Virtanen et~al\mbox{.}(2020)]%
        {2020SciPy-NMeth}
\bibfield{author}{\bibinfo{person}{Pauli Virtanen}, \bibinfo{person}{Ralf Gommers}, \bibinfo{person}{Travis~E. Oliphant}, \bibinfo{person}{Matt Haberland}, \bibinfo{person}{Tyler Reddy}, \bibinfo{person}{David Cournapeau}, \bibinfo{person}{Evgeni Burovski}, \bibinfo{person}{Pearu Peterson}, \bibinfo{person}{Warren Weckesser}, \bibinfo{person}{Jonathan Bright}, \bibinfo{person}{St{\'e}fan~J. {van der Walt}}, \bibinfo{person}{Matthew Brett}, \bibinfo{person}{Joshua Wilson}, \bibinfo{person}{K.~Jarrod Millman}, \bibinfo{person}{Nikolay Mayorov}, \bibinfo{person}{Andrew R.~J. Nelson}, \bibinfo{person}{Eric Jones}, \bibinfo{person}{Robert Kern}, \bibinfo{person}{Eric Larson}, \bibinfo{person}{C~J Carey}, \bibinfo{person}{{\.I}lhan Polat}, \bibinfo{person}{Yu Feng}, \bibinfo{person}{Eric~W. Moore}, \bibinfo{person}{Jake {VanderPlas}}, \bibinfo{person}{Denis Laxalde}, \bibinfo{person}{Josef Perktold}, \bibinfo{person}{Robert Cimrman}, \bibinfo{person}{Ian Henriksen}, \bibinfo{person}{E.~A. Quintero},
  \bibinfo{person}{Charles~R. Harris}, \bibinfo{person}{Anne~M. Archibald}, \bibinfo{person}{Ant{\^o}nio~H. Ribeiro}, \bibinfo{person}{Fabian Pedregosa}, \bibinfo{person}{Paul {van Mulbregt}}, {and} \bibinfo{person}{{SciPy 1.0 Contributors}}.} \bibinfo{year}{2020}\natexlab{}.
\newblock \showarticletitle{{{SciPy} 1.0: Fundamental Algorithms for Scientific Computing in Python}}.
\newblock \bibinfo{journal}{\emph{Nature Methods}}  \bibinfo{volume}{17} (\bibinfo{year}{2020}), \bibinfo{pages}{261--272}.
\newblock
\urldef\tempurl%
\url{https://doi.org/10.1038/s41592-019-0686-2}
\showDOI{\tempurl}


\bibitem[Wang et~al\mbox{.}(2018)]%
        {Wang_Hou_Wen_Gong_Li_Sun_Li_2018}
\bibfield{author}{\bibinfo{person}{Cheng Wang}, \bibinfo{person}{Shiwei Hou}, \bibinfo{person}{Chenglu Wen}, \bibinfo{person}{Zheng Gong}, \bibinfo{person}{Qing Li}, \bibinfo{person}{Xiaotian Sun}, {and} \bibinfo{person}{Jonathan Li}.} \bibinfo{year}{2018}\natexlab{}.
\newblock \showarticletitle{Semantic line framework-based indoor building modeling using backpacked laser scanning point cloud}.
\newblock \bibinfo{journal}{\emph{ISPRS Journal of Photogrammetry and Remote Sensing}}  \bibinfo{volume}{143} (\bibinfo{date}{Sept.} \bibinfo{year}{2018}), \bibinfo{pages}{150–166}.
\newblock
\showISSN{09242716}
\urldef\tempurl%
\url{https://doi.org/10.1016/j.isprsjprs.2018.03.025}
\showDOI{\tempurl}


\bibitem[Wang et~al\mbox{.}(2013)]%
        {Wang_Xu_Liu_Cao_Liu_Yu_Gu_2013}
\bibfield{author}{\bibinfo{person}{Jun Wang}, \bibinfo{person}{Kai Xu}, \bibinfo{person}{Ligang Liu}, \bibinfo{person}{Junjie Cao}, \bibinfo{person}{Shengjun Liu}, \bibinfo{person}{Zeyun Yu}, {and} \bibinfo{person}{Xianfeng~David Gu}.} \bibinfo{year}{2013}\natexlab{}.
\newblock \showarticletitle{Consolidation of Low-quality Point Clouds from Outdoor Scenes}.
\newblock \bibinfo{journal}{\emph{Computer Graphics Forum}} \bibinfo{volume}{32}, \bibinfo{number}{5} (\bibinfo{date}{Aug.} \bibinfo{year}{2013}), \bibinfo{pages}{207–216}.
\newblock
\showISSN{01677055}
\urldef\tempurl%
\url{https://doi.org/10.1111/cgf.12187}
\showDOI{\tempurl}


\bibitem[Zhang et~al\mbox{.}(2014)]%
        {Zhang_Huang_Zhu_Hwang_2014}
\bibfield{author}{\bibinfo{person}{Juan Zhang}, \bibinfo{person}{Wenbin Huang}, \bibinfo{person}{Xiaoqiang Zhu}, {and} \bibinfo{person}{Jenq-Neng Hwang}.} \bibinfo{year}{2014}\natexlab{}.
\newblock \showarticletitle{A subjective quality evaluation for 3D point cloud models}. In \bibinfo{booktitle}{\emph{2014 International Conference on Audio, Language and Image Processing}}. \bibinfo{publisher}{IEEE}, \bibinfo{address}{Shanghai}, \bibinfo{pages}{827–831}.
\newblock
\showISBNx{978-1-4799-3902-2}
\urldef\tempurl%
\url{https://doi.org/10.1109/ICALIP.2014.7009910}
\showDOI{\tempurl}


\bibitem[Zhou et~al\mbox{.}(2018)]%
        {Zhou2018}
\bibfield{author}{\bibinfo{person}{Qian-Yi Zhou}, \bibinfo{person}{Jaesik Park}, {and} \bibinfo{person}{Vladlen Koltun}.} \bibinfo{year}{2018}\natexlab{}.
\newblock \showarticletitle{{Open3D}: {A} Modern Library for {3D} Data Processing}.
\newblock \bibinfo{journal}{\emph{arXiv:1801.09847}} (\bibinfo{year}{2018}).
\newblock


\bibitem[Zimmer et~al\mbox{.}(2019)]%
        {Zimmer_Rangesh_Trivedi_2019}
\bibfield{author}{\bibinfo{person}{Walter Zimmer}, \bibinfo{person}{Akshay Rangesh}, {and} \bibinfo{person}{Mohan Trivedi}.} \bibinfo{year}{2019}\natexlab{}.
\newblock \showarticletitle{3D BAT: A Semi-Automatic, Web-based 3D Annotation Toolbox for Full-Surround, Multi-Modal Data Streams}.
\newblock  \bibinfo{number}{arXiv:1905.00525} (\bibinfo{date}{May} \bibinfo{year}{2019}).
\newblock
\urldef\tempurl%
\url{http://arxiv.org/abs/1905.00525}
\showURL{%
\tempurl}
\newblock
\shownote{arXiv:1905.00525 [cs]}.


\end{thebibliography}

\end{document}